# Graph Learning from Filtered Signals: Graph System and Diffusion Kernel Identification

Hilmi E. Egilmez, *Student Member, IEEE,* Eduardo Pavez, *Student Member, IEEE,* and Antonio Ortega, *Fellow, IEEE*

***Abstract*—This paper introduces a novel graph signal processing framework for building graph-based models from classes of filtered signals. In our framework, graph-based modeling is formulated as a graph system identification problem, where the goal is to learn a weighted graph (a graph Laplacian matrix) and a graph-based filter (a function of graph Laplacian matrices). In order to solve the proposed problem, an algorithm is developed to jointly identify a graph and a graph-based filter (GBF) from multiple signal/data observations. Our algorithm is valid under the assumption that GBFs are one-to-one functions. The proposed approach can be applied to learn diffusion (heat) kernels, which are popular in various fields for modeling diffusion processes. In addition, for specific choices of graph-based filters, the proposed problem reduces to a graph Laplacian estimation problem. Our experimental results demonstrate that the proposed algorithm outperforms the current state-of-the-art methods. We also implement our framework on a real climate dataset for modeling of temperature signals.**

## I. Introduction

GRAPHS are fundamental mathematical structures used in various fields to characterize data, signals and processes. Particularly in signal processing, machine learning and statistics, structured modeling of signals/data by means of graphs is essential in numerous problems including clustering [1]–[3], regularized regression and denoising [4]–[6], where graphs provide concise (sparse) representations for effective modeling and analysis of signals/data [7]. Graph signal processing [6] offers a new general paradigm for processing and analyzing signals/data on graphs, referred as *graph signals*, by using graph Laplacian matrices to extend basic signal processing operations[1] such as filtering [9], [10], transformation [11], [12] and sampling [13] on graph signals. However, in practice, datasets typically consist of an unstructured list of samples, where the graph information (representing the structural relations between samples/features) is latent. For analysis, learning, processing and algorithmic purposes, it is often useful to build graph-based models that provide a concise explanation for datasets and also reduce the dimension of the problem [14], [15] by exploiting the available prior knowledge and assumptions about the desired graph (e.g., structural information including connectivity and sparsity level) and observed data (e.g., signal smoothness).

Authors are with the Department of Electrical Engineering, University of Southern California, Los Angeles, CA, 90089 USA. Contact author e-mail: hegilmez@usc.edu. This paper was funded in part by NSF funding under grant: CCF-1410009.

[1]In [8], adjacency matrices are used to define basic processing operations. In this work, we adopt the graph Laplacian based construction in [6].

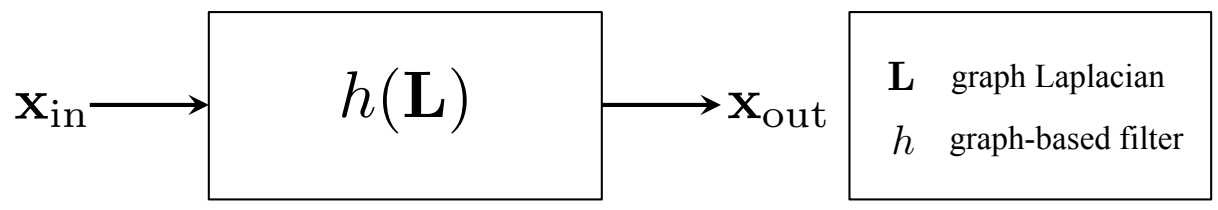

Fig. 1. Input-output relation of a graph system defined by a graph Laplacian ($\mathbf{L}$) and a graph-based filter ($h$). In this work, we focus on joint identification of $\mathbf{L}$ and $h$, given the observed data $\mathbf{x}_{\rm out}$.

The focus of this paper is to build graph-based models from signals/data, where the models of interest are defined by *graph systems* consisting of a graph (graph Laplacian matrix $\mathbf{L}$) and a graph-based filter (GBF $h$) as depicted in Fig. 1. For graph-based modeling, graph systems provide a general abstraction, in which graphs represent pairwise relations between the signal samples and GBFs model smoothness of signals, defined on a designated graph. In particular, graph systems (with specific choices of GBFs) have been introduced to represent diffusion processes. Prominent examples include diffusion kernels defined on graphs [16] and polynomials of graph Laplacian matrices used to define localized diffusion operators on graphs [6]. Models based on graph systems can be useful in a broad number of applications including signal/data processing on computer, social, sensor, energy, transportation and biological networks, when signals are observed without the knowledge of a graph and GBF associated with the underlying network. For example, signals can be diffused on an unknown network ($\mathbf{L}$) with an unknown rate ($\beta$) as the smoothness parameter of a GBF ($h_\beta$).

In the literature, there has been a lot of recent interest in the problem of learning graph-based models from data. Several papers, including [17]–[19], consider the problem of learning a graph, represented by a matrix such as graph Laplacian, where the goal is to find the graph that provides a best fit to the training data according to different criteria. A few studies (see Section III for more details) consider that the training data are the output of an unknown graph system and propose methods under different assumptions on GBFs. In [20], [21], the authors aim to learn a graph (i.e., a graph Laplacian or an adjacency matrix) from eigenvectors of the sample covariance of training data, where the GBF is an arbitrary function. In [22], the GBF is assumed to be a polynomial function of a specific order, and the goal is to learn a graph (an adjacency matrix) and coefficients of a polynomial GBF. However, with such general assumptions on GBFs, it is hard to provide guarantees about the performance of the proposed methods. Specifically in [22], the solutions require explicit sparsity conditions both on graphs and polynomials of adjacency matrices, which are restrictive and may not hold in some practical cases. Moreover, the approaches in [20], [21] are based on the assumption that the sample covariance of the observed data and the graph Laplacian have the same set of eigenvectors. Thus, their performances strictly depend on the accuracy of eigenvectors obtained from the sample covariance of data, which is not a good estimator for the true covariance of a model, especially when the number of data samples is small [23], [24]. Indeed, better eigenvector estimates can be obtained using inverse covariance estimation methods [24], [25] or graph learning methods, such as those introduced in our prior work [17], which minimize a regularized maximum-likelihood (ML) criterion.

In this paper, we propose an extension of our prior work [17] to estimate the parameters of a graph system (a graph and a GBF) from data. For this purpose, we formulate a *graph system identification (GSI)* problem with a regularized ML criterion, where our goal is to jointly identify a combinatorial graph Laplacian (CGL) matrix $\mathbf{L}$ and a GBF $h$ from multiple signal/data observations under the main assumption that the GBF $h$ is a one-to-one function. While this assumption is not as general as the assumptions on GBFs made in existing studies [20]–[22], several useful GBFs have this property, as shown in Table I. The key novelty of our work is to propose new methods that rely on the one-to-one GBF property to learn both graph and GBF with stronger optimality guarantees as compared to the approaches in [20]–[22]. In Section IV-D, we will show that this assumption is one of the sufficient conditions required for *graph system identifiability*, and it also allows our algorithm to perform a *prefiltering step* that significantly improves the estimation accuracy. Although our algorithm can be extended for any one-to-one GBFs, we focus on methods to estimate the parametric GBFs listed in Table I, which are one-to-one functions that depend on a single parameter $\beta$ and have useful applications discussed in Section IV-C. The first three GBF types in the table define basic scaling and shifting operations, and the exponential decay and $\beta$-hop localized GBFs provide diffusion-based models. Since all these GBFs yield larger filter responses ($h_\beta$) in lower graph frequencies ($\lambda$) as illustrated in Fig. 2, they can be used for modeling different classes of smooth graph signals satisfying that most of the signal energy is concentrated in the low graph frequencies.

In order to solve the GSI problem, we propose an alternating optimization algorithm that first optimizes the graph by fixing the GBF and then designs the GBF by fixing the graph. Basically, the proposed algorithm involves three main steps, which are repeated until convergence is achieved:

- The *graph-based filter (GBF) identification* step designs a GBF $h$ for current estimate of graph Laplacian $\mathbf{L}$ so that its inverse (i.e., $h^{-1}$) will be used for the prefiltering step. Note that $h$ has to be a one-to-one function for its inverse $h^{-1}$ to exist.
- The *prefiltering* step filters the observed signals using $h^{-1}$ to compute the covariance matrix that will be used in the graph Laplacian estimation step. We show that this step significantly improves the accuracy of graph estimation.
- The *graph Laplacian estimation* step learns a graph from the covariance of prefiltered signals obtained in the previous step by using the combinatorial graph Laplacian (CGL) estimation algorithm introduced in our prior work [17]. Although the present paper focuses on graphs associated with CGL matrices, our solution can be easily extended to other types of graph Laplacian matrices, e.g., generalized graph Laplacians [26].

In order to accommodate the GBFs in Table I in our algorithm, we propose specific methods (for the GBF identification step) to find the filter parameter $\beta$ fully characterizing the selected GBF. Our proposed algorithm guarantees optimal identification of $\beta$ and $\mathbf{L}$ in an $\ell_1$-regularized ML sense for frequency shifting, variance shifting and $\beta$-hop localized GBFs. However, for frequency scaling and exponential decay GBFs, our algorithm cannot find the optimal $\beta$ in general, but it guarantees that the estimated $\mathbf{L}$ is optimal up to a constant factor. In practice, the type of GBF and its parameter $\beta$ can also be selected based on the prior knowledge available about the problem and the application. In this case, different GBFs and their parameters can be tested until the estimated graph and GBF pair achieves the desired performance (e.g., in terms of mean square error, likelihood or sparsity) where the parameter $\beta$ serves as a regularization parameter, which can be used for tuning smoothness of the signal, for example. Moreover, our algorithm can be extended to support GBFs beyond the filters in Table I (including one-to-one functions with more than one parameter) by developing specific methods for the GBF identification step. As long as a specified GBF ($h$) has an inverse function ($h^{-1}$), the proposed prefiltering and graph Laplacian estimation steps can be directly utilized to learn graphs from signals/data.

To the best of our knowledge, this is the first work that (i) formulates the graph-based modeling problem as identification of graph systems with the types of GBFs in Table I under a regularized ML criterion and (ii) proposes a prefiltering-based algorithm to jointly identify a graph and a GBF. Existing related studies (see Section III) consider optimization of different criteria (see Table III), and do not use prefiltering, which can be shown to be optimal in some cases (see Section V). The proposed approach can significantly improve the accuracy of *graph learning from filtered signals*, as compared to the existing methods that estimate a graph directly from observed signals without prefiltering. Particularly, if observed signals are diffused/filtered on an unknown network/graph to be learned, applying a graph learning algorithm (e.g., CGL algorithm in [17]) on diffused signals potentially results in a dense graph due to diffusion, even when the underlying graph is sparse. On the other hand, our proposed algorithm reverses the effect of diffusion via prefiltering ($h^{-1}$), whose corresponding GBF ($h$) is jointly estimated with the graph ($\mathbf{L}$). Thus, the latent graph can be learned more accurately. Note that the diffusion (heat) kernels [16] used in a number of applications [4], [5], [27] are special cases of graph systems with exponential decay GBFs, and thus our proposed algorithm can be applied to learn their parameters.

The rest of the paper is organized as follows. Section II presents the notation and some concepts used throughout the paper. Section III discusses the prior related work. In Section IV, the GSI problem and its variations are formulated. Additionally, the graph system identifiability conditions are introduced. In Section V, we derive optimality conditions and develop methods to solve the GSI problem. The experimental results are presented in Section VI, and Section VII draws some conclusions.

TABLE I
GBFs WITH PARAMETER $\beta$ AND CORRESPONDING INVERSE FUNCTIONS WHERE $\lambda^{\dagger}$ DENOTES THE PSEUDOINVERSE OF A SCALAR $\lambda$, THAT IS, $\lambda^{\dagger} = 1/\lambda$ FOR $\lambda \neq 0$ AND $\lambda^{\dagger} = 0$ FOR $\lambda = 0$.

| Filter Name | $h_\beta(\lambda)$ | $h_\beta^{-1}(s)$ | Range of $\beta$ |
|---|---|---|---|
| Frequency scaling | $\begin{cases} \frac{1}{\beta\lambda} & \lambda > 0 \\ 0 & \lambda = 0 \end{cases}$ | $\begin{cases} \frac{1}{\beta s} & s > 0 \\ 0 & s = 0 \end{cases}$ | $\beta \in \mathbb{R}, \beta > 0$ |
| Frequency shifting | $(\lambda + \beta)^{\dagger}$ | $\frac{1}{s} - \beta$ | $\beta \in \mathbb{R}, \beta \geq 0$ |
| Variance shifting | $\lambda^{\dagger} + \beta$ | $(s - \beta)^{\dagger}$ | $\beta \in \mathbb{R}, \beta \geq 0$ |
| Exponential decay | $\exp(-\beta\lambda)$ | $-\log(s)/\beta$ | $\beta \in \mathbb{R}, \beta > 0$ |
| $\beta$-hop localized | $\begin{cases} \frac{1}{\lambda^\beta} & \lambda > 0 \\ 0 & \lambda = 0 \end{cases}$ | $\begin{cases} \left(\frac{1}{s}\right)^{\frac{1}{\beta}} & s > 0 \\ 0 & s = 0 \end{cases}$ | $\beta \in \mathbb{N}, \beta \geq 1$ |

TABLE II
LIST OF SYMBOLS AND THEIR MEANING

| Symbols | Meaning |
|---|---|
| $\mathbf{L} \mid \mathcal{L}, \mathcal{L}_c$ | graph Laplacian or CGL matrix $\mid$ set of CGLs |
| $h, h_\beta \mid \mathcal{H}$ | GBF function $\mid$ set of GBFs |
| $\lambda_i, \lambda_i(\mathbf{L})$ | $i$-th eigenvalue of $\mathbf{L}$ in ascending order |
| $n \mid k$ | number of vertices $\mid$ number of data samples |
| $\mathbf{I} \mid \mathbf{W} \mid \mathbf{D}$ | identity matrix $\mid$ adjacency matrix $\mid$ degree matrix |
| $\mathbf{1} \mid \mathbf{0}$ | column vector of ones $\mid$ column vector of zeros |
| $\mathbf{\Theta}^{-1} \mid \mathbf{\Theta}^{\dagger}$ | inverse of $\mathbf{\Theta}$ $\mid$ pseudo-inverse of $\mathbf{\Theta}$ |
| $\mathbf{\Theta}^{\mathsf{T}} \mid \boldsymbol{\theta}^{\mathsf{T}}$ | transpose of $\mathbf{\Theta}$ $\mid$ transpose of $\boldsymbol{\theta}$ |
| $\det(\mathbf{\Theta}) \mid \lvert\mathbf{\Theta}\rvert$ | determinant of $\mathbf{\Theta}$ $\mid$ pseudo-determinant of $\mathbf{\Theta}$ |
| $(\mathbf{\Theta})_{ij}$ | entry of $\mathbf{\Theta}$ at $i$-th row and $j$-th column |
| $(\boldsymbol{\theta})_i$ | $i$-th entry of $\boldsymbol{\theta}$ |
| $\geq (\leq)$ | element-wise greater (less) than or equal to operator |
| $\mathbf{\Theta} \succeq 0$ | $\mathbf{\Theta}$ is a positive semidefinite matrix |
| $\mathrm{Tr} \mid \mathrm{logdet}(\mathbf{\Theta})$ | trace operator $\mid$ natural logarithm of $\det(\mathbf{\Theta})$ |
| $\mathsf{p}(\mathbf{x})$ | probability density function of random vector $\mathbf{x}$ |
| $\mathbf{x} \sim \mathsf{N}(\mathbf{0}, \mathbf{\Sigma})$ | zero-mean multivariate Gaussian with covariance $\mathbf{\Sigma}$ |
| $\lVert\boldsymbol{\theta}\rVert_1, \lVert\mathbf{\Theta}\rVert_1$ | sum of absolute values of all elements ($\ell_1$-norm) |
| $\lVert\boldsymbol{\theta}\rVert_2^2, \lVert\mathbf{\Theta}\rVert_F^2$ | sum of squared values of all elements |

## II. NOTATION AND PRELIMINARIES

Throughout the paper, lowercase normal (e.g., $a$ and $\theta$), lowercase bold (e.g., $\mathbf{a}$ and $\boldsymbol{\theta}$) and uppercase bold (e.g., $\mathbf{A}$ and $\mathbf{\Theta}$) letters denote scalars, vectors and matrices, respectively. Unless otherwise stated, calligraphic capital letters (e.g., $\mathcal{E}$ and $\mathcal{S}$) represent sets. Notation is summarized in Table II.

The graph-based models considered in this paper are defined based on undirected, simple weighted graphs with nonnegative edge weights and no self-loops. Let $\mathcal{G} = (\mathcal{V}, \mathcal{E}, f_w)$ be a simple weighted graph with $n$ vertices in the set $\mathcal{V} = \{v_1, \dots, v_n\}$, where the edge set $\mathcal{E} = \{\, e \mid f_w(e) \neq 0, \forall\, e \in \mathcal{P}_u\}$ is a subset of $\mathcal{P}_u$, the set of all unordered pairs of distinct vertices, and $f_w((v_i, v_j)) \geq 0$ for $i \neq j$ is a real-valued edge weight function. The *adjacency matrix* of $\mathcal{G}$ is an $n \times n$ symmetric matrix, $\mathbf{W}$, such that $(\mathbf{W})_{ij} = (\mathbf{W})_{ji} = f_w((v_i, v_j))$ for $(v_i, v_j) \in \mathcal{P}_u$. The *degree matrix* of $\mathcal{G}$ is an $n \times n$ diagonal matrix, $\mathbf{D}$, with entries $(\mathbf{D})_{ii} = \sum_{j=1}^{n} (\mathbf{W})_{ij}$ and $(\mathbf{D})_{ij} = 0$ for $i \neq j$. The *combinatorial graph Laplacian (CGL)* of $\mathcal{G}$ is defined as $\mathbf{L} = \mathbf{D} - \mathbf{W}$. Alternatively, the set of CGL matrices can also be written as

$$\mathcal{L}_c = \{\mathbf{L} \mid \mathbf{L} \succeq 0,\ (\mathbf{L})_{ij} \leq 0 \text{ for } i \neq j,\ \mathbf{L}\mathbf{1} = \mathbf{0}\}\,. \quad (1)$$

By construction, CGLs are symmetric and positive semidefinite matrices, so each of them has a complete set of orthogonal eigenvectors $\mathbf{u}_1, \mathbf{u}_2, \dots, \mathbf{u}_n$ whose associated eigenvalues $\lambda_1 \leq \lambda_2 \leq \cdots \leq \lambda_n$ are nonnegative real numbers[2]. In addition, the CGL of a connected graph always has a zero eigenvalue (i.e., $\lambda_1 = 0$ with multiplicity one) whose associated eigenvector is $\mathbf{u}_1 = (1/\sqrt{n})\mathbf{1}$.

In graph signal processing, the eigenpairs of a CGL matrix, $(\lambda_1, \mathbf{u}_1), (\lambda_2, \mathbf{u}_2), \dots, (\lambda_n, \mathbf{u}_n)$, provide a Fourier-like spectral interpretation for signals defined on graphs, where the graph frequency spectrum is defined by eigenvalues of the CGL, which are called graph frequencies, and eigenvectors of the CGL $\mathbf{u}_1, \mathbf{u}_2, \dots, \mathbf{u}_n$ are the harmonics associated with the graph frequencies. Based on the eigenvectors of a CGL matrix $\mathbf{L}$, the *graph Fourier transform* (GFT) is defined as the orthogonal matrix $\mathbf{U}$ (i.e., satisfying $\mathbf{U}^{\mathsf{T}}\mathbf{U} = \mathbf{I}$) obtained by eigendecomposition of $\mathbf{L} = \mathbf{U}\mathbf{\Lambda}\mathbf{U}^{\mathsf{T}}$, where $\mathbf{\Lambda}$ denotes the diagonal matrix consisting of eigenvalues $\lambda_1, \lambda_2, \dots, \lambda_n$ (graph frequencies).

For a given signal $\mathbf{x} = [x_1\, x_2\, \cdots\, x_n]^{\mathsf{T}}$ defined on a graph $\mathcal{G}$ with $n$ vertices, where $x_i$ is attached to $v_i$ ($i$-th vertex), the GFT of $\mathbf{x}$ is $\widehat{\mathbf{x}} = \mathbf{U}^{\mathsf{T}}\mathbf{x}$. Naturally, the variation of the GFT basis vectors gradually increase with the graph frequencies, and the GBF basis vectors corresponding to low frequencies are relatively smooth. As a specific example, the GFT basis vector $\mathbf{u}_1 = (1/\sqrt{n})\mathbf{1}$, associated with lowest graph frequency ($\lambda_1 = 0$), is the smoothest among other GFT basis vectors. To quantify smoothness of a graph signal $\mathbf{x}$, a common variation measure used in graph signal processing is the *graph Laplacian quadratic form*, $\mathbf{x}^{\mathsf{T}}\mathbf{L}\mathbf{x}$, which can be written in terms of edge weights of $\mathcal{G}$ as

$$\mathbf{x}^{\mathsf{T}}\mathbf{L}\mathbf{x} = \sum_{(i,j) \in \mathcal{I}} (\mathbf{W})_{ij}\,(x_i - x_j)^2 \quad (2)$$

where $(\mathbf{W})_{ij} = -(\mathbf{L})_{ij}$, and $\mathcal{I} = \{(i,j) \mid (v_i, v_j) \in \mathcal{E}\}$ is the set of index pairs of vertices associated with the edge set $\mathcal{E}$. A smaller Laplacian quadratic term ($\mathbf{x}^{\mathsf{T}}\mathbf{L}\mathbf{x}$) indicates a smoother graph signal ($\mathbf{x}$). A more general notion of smoothness can be obtained by defining *graph-based filters* (GBFs), i.e., matrix functions $h(\mathbf{L}) = \mathbf{U}h(\mathbf{\Lambda})\mathbf{U}^{\mathsf{T}}$, where $\mathbf{U}$ is the GFT and $(h(\mathbf{\Lambda}))_{ii} = h((\mathbf{\Lambda})_{ii}) = h(\lambda_i)\ \forall i$. Note that a GBF $h$ serves as a scalar function of $\lambda$ that maps graph frequencies $\lambda_1, \dots, \lambda_n$

[2]Without loss of generality, we assume that the eigenvalues are ordered as $\lambda_1 \leq \lambda_2 \leq \cdots \leq \lambda_n$ throughout the paper.

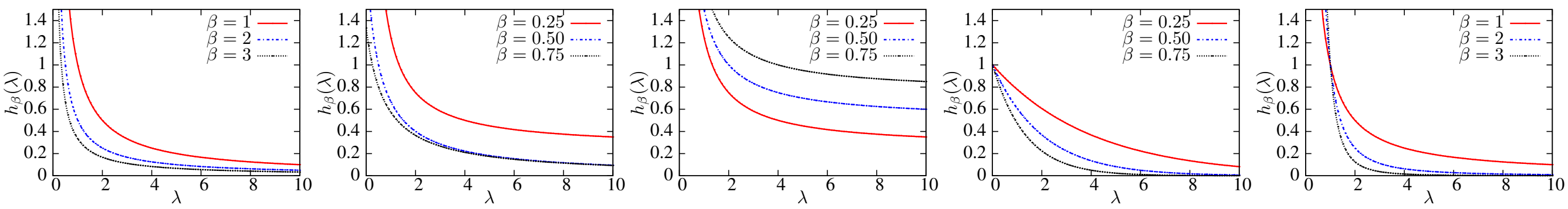

(a) Frequency scaling GBF (b) Frequency shifting GBF (c) Variance shifting GBF (d) Exponential decay GBF (e) $\beta$-hop localized GBF

Fig. 2. The illustration of graph-based filters (GBFs) in Table I with different $\beta$ parameters as a function of graph frequency $h_\beta(\lambda)$.

to filter responses[3] $h(\lambda_1), \dots, h(\lambda_n)$. Thus, the measure in (2) can be generalized using a GBF $h$ as

$$\mathbf{x}^\intercal h(\mathbf{L})\mathbf{x} = \mathbf{x}^\intercal \mathbf{U} h(\mathbf{\Lambda})\mathbf{U}^\intercal \mathbf{x} = \sum_{i=1}^{n} h(\lambda_i)\widehat{x}_i^2, \tag{3}$$

where $\widehat{x}_i = \mathbf{u}_i^\intercal \mathbf{x}$ for $i = 1, \dots, n$ denote GFT coefficients.

Based on a graph and a GBF, we formally define a graph system as follows.

**Definition 1** (Graph System)**.** A graph system is defined by a simple graph $\mathcal{G}$ and a GBF $h$ such that the input-output relation of the system is $\mathbf{x}_{\text{out}} = h(\mathbf{L})\mathbf{x}_{\text{in}}$, where $\mathbf{L} \in \mathcal{L}_c$ is a CGL matrix associated with $\mathcal{G}$, and $\mathbf{x}_{\text{in}}$ is the input signal vector.

In this paper, the graph system parameter $h$ is selected from a set of GBFs $\mathcal{H}$ determined based on the GBF types in Table I. As a specific example, the exponential decay GBFs lead to the set $\mathcal{H} = \{h \,|\, h(\lambda) = \exp(-\beta\lambda),\ \beta \in \mathbb{R} \text{ and } \beta > 0\}$ and the operator $h(\mathbf{L}) = \exp(-\beta\mathbf{L})$, which is also known as the diffusion (heat) kernel defined on a graph [16].

**Definition 2** (Diffusion Kernels on Graphs)**.** The diffusion kernel over graph $\mathcal{G}$ is the matrix exponential of $\mathbf{L}$, that is

$$\exp(-\beta\mathbf{L}) = \lim_{t\to\infty}\left(\mathbf{I} - \frac{\beta\mathbf{L}}{t}\right)^t \quad t \in \mathbb{N}, \tag{4}$$

where $\mathbf{L}$ denotes a graph Laplacian associated with $\mathcal{G}$ and the parameter $\beta$ is a real number called diffusion bandwidth.

Appendix A presents a derivation of diffusion kernels (i.e., exponential GBFs) based on a basic class of random diffusion processes defined by CGLs.

## III. Related Work

Table III summarizes prior related studies by comparing against our present work in terms of (i) target variables, (ii) underlying assumptions and (iii) optimization criteria. In our previous work on graph learning from data [17], we have proposed algorithms for estimating models based on three different types of graph Laplacians (including CGLs). The present paper extends our prior work by introducing more general graph-based models based on graph systems, defined by a CGL and a GBF, and proposing an algorithm to learn parameters of a graph system from multiple signal/data observations. In the literature, there are several papers on graph learning from signals/data[4]. Methods to estimate CGLs from smooth signals are proposed in [18] and [19], but, in contrast with our work, GBFs are not considered explicitly. Instead, the graph estimation problem is formulated as minimization of a regularized graph Laplacian quadratic form (i.e., a smoothness metric for graph signals). In [20], [21], the authors focus on learning graph shift/diffusion operators (e.g., adjacency and graph Laplacian matrices) from a complete set of eigenvectors that has to be computed from observed data. More specifically, Segarra *et al.* [20] solve a sparse recovery problem by minimizing the $\ell_1$-norm of the target variable (i.e., minimizing $\|\mathbf{\Theta}\|_1$ where $\mathbf{\Theta}$ is the target variable) to infer the topology of a graph, and Pasdeloup *et al.* [21] estimate an adjacency matrix by minimizing its trace, $\mathrm{Tr}(\mathbf{\Theta})$, as well as its $\ell_1$-norm, $\|\mathbf{\Theta}\|_1$. In fact, the problems in [20] and [21] are equivalent if the target matrix is constrained to be a CGL, since the problems of minimizing $\|\mathbf{L}\|_1$ and $\mathrm{Tr}(\mathbf{L})$ over the set of CGL matrices (i.e., $\mathbf{L} \in \mathcal{L}_c$) lead to the same solution. Note that both of these methods [20], [21] only use the eigenvectors of the sample covariance, while its eigenvalues (which also carry graph information implicitly) are not exploited in graph estimation. Although our approach also requires a set eigenvectors to be computed, they are not directly used to estimate a graph. Instead, the computed eigenvectors and their associated eigenvalues are used in the prefiltering step, then a graph is estimated from the covariance of prefiltered signals by minimizing a regularized ML criterion. The optimality of our prefiltering-based approach is discussed in Section V.

Thanou *et al.* [28] address the estimation of a graph and a sparse input (i.e., localized sources) from a set of observed signals, and they propose a dictionary-based method, where a graph estimation problem is solved to construct a dictionary consisting of multiple diffusion kernels and the resulting dictionary is used for identifying the localized sources in the diffusion process. Noting that our paper focuses on the GSI problem without locality assumptions on diffused/filtered signals (see Table III), when no locality assumptions are imposed and a single diffusion kernel is used in the dictionary, the problem in [28] reduces to the graph estimation problem in [18], formulated as minimization of a regularized graph Laplacian quadratic form. In contrast with the work in [28], our algorithm iteratively solves a different graph learning problem with a regularized ML criterion (i.e., the CGL estimation problem in [17]) and also performs prefiltering on observed signals. Since our algorithm often results in a more accurate graph estimation as compared to the method in [18] (see Section VI for the details), it can be extended to construct potentially better dictionaries as alternatives to the ones in [28] used for identifying localized sources. Such an extension is out of the scope of the present paper. In [22], Mei and Moura address the estimation of polynomials of adjacency matrices by solving a regularized least-squares problem. As their counterparts, polynomials of graph Laplacians can be used to approximate the GBFs in Table I as well as many other types of filters such as bandlimited GBFs. Since polynomial filters provide more degrees of freedom in designing GBFs, they can be considered as more general compared to our GBFs of interest. However, polynomials are not one-to-one functions in general, so our proposed prefiltering step cannot be applied. The algorithm in [22] provides some optimality guarantees in a mean-square sense, but this is under a restrictive set of assumptions, which require the polynomials of adjacency matrices to be sparse[5]. Our proposed algorithm provides stronger optimality guarantees in a regularized ML sense without imposing any assumptions on the sparsity of graphs (see Table III), as long as GBFs are one-to-one. The identification of graph systems with polynomial GBFs will be studied as part of our future work.

[3]Filter responses corresponding to the eigenvalues with multiplicity more than one have the same value. Formally, if $\lambda_i = \lambda_{i+1} = \cdots = \lambda_{i+c-1}$ for some $i > 1$ and multiplicity $c > 1$, then $h(\lambda_i) = h(\lambda_{i+1}) = \cdots = h(\lambda_{i+c-1})$.

[4]We refer to [17] for a discussion of methods beyond CGL estimation.

TABLE III
AN OVERVIEW OF THE RELATED WORK ON LEARNING GRAPHS FROM SMOOTH/DIFFUSED SIGNALS.

| References | Target Variable(s) | Assumptions | | | Optimization Criterion |
|---|---|---|---|---|---|
| | | Graph | Filter | Signal | |
| Dong *et al.* [18]<br>Kalofolias [19] | graph Laplacian | None | None | None | regularized Laplacian quadratic form |
| Segarra *et al.* [20] | graph shift operator | None | None | None | $\ell_1$-norm of graph shift operator |
| Pasdeloup *et al.* [21] | adjacency matrix | None | None | None | trace and $\ell_1$-norm of adjacency |
| Thanou *et al.* [28] | multiple heat kernels<br>source locations | None | Heat kernel | Localized | regularized least squares |
| Mei and Moura [22] | polynomials of adjacency | Sparse | Polynomial | None | regularized least squares |
| Egilmez *et al.* [17] | graph Laplacian | None | None | None | regularized maximum likelihood |
| This work | graph Laplacian<br>GBF (Table I) | None | One-to-one function | None | regularized maximum likelihood |

## IV. Problem Formulation: Graph System Identification

Our formulation is based on the following general assumption on a GBF $h$, which holds for the GBFs in Table I.

**Assumption 1.** We assume that a graph-based filter $h(\lambda)$ is a nonnegative and one-to-one function of $\lambda$.

### A. Filtered Signal Model

We formulate the GSI problem in a probabilistic setting by assuming that the observed (filtered) signals have been sampled from a zero mean $n$-variate Gaussian distribution[6]

$$\mathbf{x} \sim \mathsf{N}(\mathbf{0}, h(\mathbf{L})),$$

$$\mathsf{p}(\mathbf{x}|h(\mathbf{L})) = \frac{1}{(2\pi)^{n/2}|h(\mathbf{L})|^{1/2}} \exp\left(-\frac{1}{2}\mathbf{x}^{\mathsf{T}} h(\mathbf{L})^{\dagger}\mathbf{x}\right), \quad (5)$$

with the covariance $\mathbf{\Sigma} = h(\mathbf{L})$ defined based on a CGL matrix $\mathbf{L}$ and a GBF $h$.

For modeling smooth graph signals, it is reasonable to choose $h(\lambda)$ to be a monotonically decreasing function[7] satisfying $h(\lambda_1) \geq h(\lambda_2) \geq \cdots \geq h(\lambda_{n-1}) \geq h(\lambda_n) > 0$, where the graph frequencies (eigenvalues of $\mathbf{L}$) are ordered as $0 = \lambda_1 \leq \lambda_2 \leq \cdots \leq \lambda_n$. Thus, the corresponding covariance matrix $h(\mathbf{L})$ represents graph signals whose energy is larger in lower graph frequencies. Note that the nonnegativity condition in Assumption 1, that is, $h(\lambda_i) \geq 0$ for $i = 1, \ldots, n$, ensures that the covariance $h(\mathbf{L})$ is a positive semidefinite matrix.

From the graph system perspective (see Fig. 1 and Definition 1), the above probabilistic model corresponds to the case when the input is the $n$-variate white Gaussian noise $\mathbf{x}_{\text{in}} \sim \mathsf{N}(\mathbf{0}, \mathbf{\Sigma}_{\text{in}} = \mathbf{I})$. Then, the covariance of the output vector $\mathbf{x}_{\text{out}} = h(\mathbf{L})\mathbf{x}_{\text{in}}$ becomes $\mathbf{\Sigma}_{\text{out}} = h(\mathbf{L})\mathbf{\Sigma}_{\text{in}} h(\mathbf{L})^{\mathsf{T}} = h(\mathbf{L})^2$, which can be translated into our model in (5) by simply transforming $\mathbf{\Sigma}_{\text{out}}$ as $\mathbf{\Sigma} = \mathbf{\Sigma}_{\text{out}}^{1/2} = h(\mathbf{L})$, where the mapping between $h(\mathbf{L})^2$ and $h(\mathbf{L})$ is one-to-one (i.e., there is no loss of information after transformation), since eigenvalues of $h(\mathbf{L})$ are nonnegative.

### B. Problem Formulation

Our goal is to estimate the graph system parameters $\mathbf{L}$ and $h$ based on $k$ random samples, $\mathbf{x}_i$ for $i = 1, \ldots, k$, obtained from the signal model in (5) by maximizing the likelihood of $h(\mathbf{L})$, that is

$$\prod_{i=1}^{k} \mathsf{p}(\mathbf{x}_i|h(\mathbf{L})) = c\,|h(\mathbf{L})|^{-\frac{k}{2}} \prod_{i=1}^{k} \exp\left(-\frac{1}{2}\mathbf{x}_i{}^{\mathsf{T}} h(\mathbf{L})^{\dagger}\mathbf{x}_i\right) \quad (6)$$

with a constant $c = (2\pi)^{-\frac{kn}{2}}$. The maximization of (6) can be equivalently stated as minimizing its negative log-likelihood,

[5] In practice, powers of adjacency/Laplacian matrices lead to dense matrices.

[6] The zero mean assumption is made to simplify the notation. Our model can be trivially extended to a multivariate Gaussian distributions with nonzero mean.

[7] All of the GBFs in Table I are monotonically decreasing functions on the interval $\lambda \in (0, \infty)$. The exponential decay and frequency shifting GBFs further satisfies $h(\lambda_1) \geq h(\lambda_2)$ at the zero frequency (i.e., $0 = \lambda_1 \leq \lambda_2$), while for the other GBFs, we have $h(\lambda_2) \geq h(\lambda_1)$.

which leads to the following problem for estimating the graph system parameters $\mathbf{L}$ and $h$,

$$(\widehat{\mathbf{L}},\widehat{h}) = \underset{\mathbf{L},h}{\operatorname{argmin}} \left\{ \frac{1}{2}\sum_{i=1}^{k} \operatorname{Tr}\left(\mathbf{x}_i{}^{\mathsf{T}} h(\mathbf{L})^{\dagger}\mathbf{x}_i\right) - \frac{k}{2}\log|h(\mathbf{L})^{\dagger}| \right\} = \underset{\mathbf{L},h}{\operatorname{argmin}} \left\{ \operatorname{Tr}\left(h(\mathbf{L})^{\dagger}\mathbf{S}\right) - \log|h(\mathbf{L})^{\dagger}| \right\} \tag{7}$$

where $\mathbf{S}$ denotes the sample covariance calculated using $k$ samples, $\mathbf{x}_i$ for $i = 1, 2, \dots, k$. In our problem formulation, we additionally introduce a sparsity promoting weighted $\ell_1$-regularization, $\|\mathbf{L} \odot \mathbf{H}\|_1$ for robust graph estimation, where $\mathbf{H}$ is symmetric regularization matrix and $\odot$ denotes element-wise multiplication. Thus, the proposed GSI problem is formulated as follows:

$$\begin{aligned} \underset{\mathbf{L}\succeq 0,\beta}{\text{minimize}} \quad & \operatorname{Tr}(h_\beta(\mathbf{L})^{\dagger}\mathbf{S}) - \log|h_\beta(\mathbf{L})^{\dagger}| + \|\mathbf{L} \odot \mathbf{H}\|_1 \\ \text{subject to} \quad & \mathbf{L}\,\mathbf{1} = \mathbf{0}, \quad (\mathbf{L})_{ij} \le 0 \quad i \ne j \end{aligned} \tag{8}$$

where $\beta$ is the (unknown) parameter for a given type of GBF $h_\beta$ from Table I, and $\mathbf{H}$ denotes the regularization matrix. The constraints ensure that $\mathbf{L}$ is a CGL matrix[8]. In this work, we particularly choose $\mathbf{H} = \alpha(2\mathbf{I} - \mathbf{1}\mathbf{1}^{\mathsf{T}})$ so that the regularization term in (8) reduces to the standard $\ell_1$-regularization $\alpha\|\mathbf{L}\|_1$ with parameter $\alpha$.

### *C. Special Cases of Graph System Identification Problem and Applications of Graph-based Filters*

**Graph Learning.** The problem in (8) can be reduced to the CGL problem proposed in our previous work [17] by choosing the filter $h_\beta$ to be the frequency scaling or $\beta$-hop localized GBFs with $\beta = 1$, so that we have

$$h_\beta(\lambda) = \lambda^{\dagger} = \begin{cases} 1/\lambda & \lambda > 0 \\ 0 & \lambda = 0 \end{cases} \text{ that is } h_\beta(\mathbf{L}) = \mathbf{L}^{\dagger}. \tag{9}$$

Since $h_\beta(\mathbf{L})^{\dagger} = \mathbf{L}$, we can rewrite the objective function in (8) as

$$\operatorname{Tr}(\mathbf{L}\mathbf{S}) - \log|\mathbf{L}| + \|\mathbf{L} \odot \mathbf{H}\|_1 \tag{10}$$

which is the objective function of graph learning problems originally introduced in [17].

**Graph Learning from Noisy Signals.** The variance shifting filter corresponds to a noisy signal model with variance $\beta = \sigma^2$. Specifically, assuming that the noisy signal is modeled as $\widehat{\mathbf{x}} = \mathbf{x} + \mathbf{e}$, where $\mathbf{x} \sim \mathsf{N}(\mathbf{0}, \mathbf{L}^{\dagger})$ denotes the original signal, and $\mathbf{e} \sim \mathsf{N}(\mathbf{0}, \sigma^2\mathbf{I})$ is the additive white Gaussian noise vector independent of $\mathbf{x}$ with variance $\sigma^2$, then the noisy signal $\widehat{\mathbf{x}}$ is distributed as $\widehat{\mathbf{x}} \sim \mathsf{N}(\mathbf{0}, \widehat{\mathbf{\Sigma}} = \mathbf{L}^{\dagger} + \sigma^2\mathbf{I})$. The same model is obtained by using a variance shifting filter with $\beta = \sigma^2$ so that

$$h_\beta(\lambda) = \lambda^{\dagger} + \beta \text{ and } h_\beta(\mathbf{L}) = \mathbf{L}^{\dagger} + \beta\mathbf{I} = \mathbf{L}^{\dagger} + \sigma^2\mathbf{I}. \tag{11}$$

With this type of GBFs, our GSI problem in (8) can be solved to learn graphs from noisy signals by identifying $\mathbf{L}$ and the noise parameter $\beta = \sigma^2$. This problem can also be viewed as an extension of our formulations in [17] derived based on signal models that are assumed to be noise-free (i.e., $\beta = 0$), so it can be solved to improve the performance of graph learning from data in the presence of noise [29].

**Graph Learning from Frequency Shifted Signals.** For the shifted frequency filter with parameter $\beta$, we have

$$h_\beta(\lambda) = (\lambda + \beta)^{\dagger} \text{ and } h_\beta(\mathbf{L}) = (\mathbf{L} + \beta\mathbf{I})^{\dagger}. \tag{12}$$

By substituting $h_\beta(\mathbf{L})$ with $(\mathbf{L} + \beta\mathbf{I})^{\dagger}$, the problem in (8) can be written as

$$\begin{aligned} \underset{\widetilde{\mathbf{L}}\succeq 0,\beta\ge 0}{\text{minimize}} \quad & \operatorname{Tr}(\widetilde{\mathbf{L}}\mathbf{S}) - \log|\widetilde{\mathbf{L}}| + \|\widetilde{\mathbf{L}} \odot \mathbf{H}\|_1 \\ \text{subject to} \quad & \widetilde{\mathbf{L}} = \mathbf{L} + \beta\mathbf{I}, \quad \mathbf{L}\mathbf{1} = \mathbf{0}, \quad (\mathbf{L})_{ij} \le 0 \quad i \ne j \end{aligned} \tag{13}$$

which is the shifted combinatorial Laplacian (SCGL) estimation problem that is originally proposed in [30].

**Diffusion (Heat) Kernel Learning.** To learn diffusion kernels, formally defined in Definition 2, the proposed GSI problem in (8) can be modified by choosing $h_\beta(\lambda)$ as the exponential decay filter, so we get

$$h_\beta(\mathbf{L}) = \mathbf{U}h_\beta(\mathbf{\Lambda})\mathbf{U}^{\mathsf{T}} = \mathbf{U}\exp(-\beta\mathbf{\Lambda})\mathbf{U}^{\mathsf{T}} = \exp(-\beta\mathbf{L}). \tag{14}$$

The resulting problem can be solved to learn diffusion kernels from signals/data, which are popular in many applications [4], [5], [27].

**Graph Learning from $\beta$-hop Localized Signals.** To learn graphs from $\beta$-hop localized signals, where $\beta$ is a positive integer, the GBF can be selected as $h(\lambda) = (\lambda^{\dagger})^{\beta}$ so that the corresponding inverse covariance (i.e., precision) matrix in (8) is $h(\mathbf{L})^{\dagger} = \mathbf{L}^{\beta}$, which is generally not a graph Laplacian matrix due to the exponent $\beta$. However, it defines diffusion operators (on a graph associated with $\mathbf{L}$) that can be used as alternatives to heat kernels, and the corresponding signal model leads to $\beta$-hop localized/diffused signals, in which each sample depends on samples located within the neighborhood at most $\beta$-hops away.

### *D. Graph System Identifiability*

In statistical learning theory, a probabilistic model is *identifiable* if it is possible to learn the model parameters exactly from infinite number of data samples. This requires different model parameters to generate different probability distributions [31], which is formally stated in the following definition.

**Definition 3** (Model Identifiability [31])**.** Let $\mathcal{M} = \{\mathsf{p}(\mathbf{x}|\mathbf{\Theta}) : \mathbf{\Theta} \in \mathcal{P}_{\mathbf{\Theta}}\}$ be the family of probabilistic models defined by the parameter space $\mathcal{P}_{\mathbf{\Theta}}$. $\mathcal{M}$ is identifiable if the mapping from $\mathcal{P}_{\mathbf{\Theta}}$ to $\mathcal{M}$ is one-to-one, that is, if $\mathbf{\Theta}_1 \ne \mathbf{\Theta}_2$ then $\mathsf{p}(\mathbf{x}|\mathbf{\Theta}_1) \ne \mathsf{p}(\mathbf{x}|\mathbf{\Theta}_2)$ for arbitrary $\mathbf{\Theta}_1, \mathbf{\Theta}_2 \in \mathcal{P}_{\mathbf{\Theta}}$.

Based on the above definition, we introduce three different notions of identifiability for graph systems characterizing the probabilistic model discussed in Section IV-A.

**Definition 4** (Notions of Identifiability)**.** Let the parameter set of the signal model in (5) be $\mathcal{P}_{\mathbf{\Theta}} = \{(\mathbf{L}, h) \,|\, \mathbf{L} \in \mathcal{L}, h \in \mathcal{H}\}$ such that a *class of graph systems* is defined by a specific choice of sets $\mathcal{L}$ and $\mathcal{H}$, denoting admissible graphs and filters in the class. For any such class of graph systems, we define the following types of identifiability:

- A class is *filter identifiable* if $h_1 \ne h_2$ implies $h_1(\mathbf{L}) \ne h_2(\mathbf{L})$ for arbitrary $\mathbf{L} \in \mathcal{L}$ and $h_1, h_2 \in \mathcal{H}$.

[8]The formulation can be trivially extended for different types of graph Laplacians (e.g., generalized graph Laplacian) discussed in [17].

- A class is *graph identifiable* if $\mathbf{L}_1 \neq \mathbf{L}_2$ implies $h(\mathbf{L}_1) \neq h(\mathbf{L}_2)$ for arbitrary $h \in \mathcal{H}$ and $\mathbf{L}_1, \mathbf{L}_2 \in \mathcal{L}$.
- A class is *jointly identifiable* if any two pairs satisfying $(\mathbf{L}_1, h_1) \neq (\mathbf{L}_2, h_2)$ lead to $h_1(\mathbf{L}_1) \neq h_2(\mathbf{L}_2)$ for arbitrary $\mathbf{L}_1, \mathbf{L}_2 \in \mathcal{L}$ and $h_1, h_2 \in \mathcal{H}$.

The following proposition categorizes the classes of systems defined based on parametric GBFs in Table I in terms of their identifiability.

**Proposition 1.** *Let $\mathcal{L}_c$ and $\mathcal{H}$ define a class of graph systems. All classes corresponding to the parametric GBFs in Table I are both filter and graph identifiable. In addition, the classes with $\beta$-hop localized, frequency shifting or variance shifting GBFs in Table I are jointly identifiable.*

*Proof:* The proof follows from Definition 4 by checking types of identifiability for each GBF with different $\beta$ parameters. For filter identifiability the proof is straightforward, because two different parameters $\beta_1 \neq \beta_2$ imply $h_{\beta_1}(\mathbf{L}) \neq h_{\beta_2}(\mathbf{L})$. Since GBFs of interest are one-to-one functions, corresponding classes are also graph identifiable by Proposition 2. The classes with frequency scaling and exponential decay GBFs are not jointly identifiable, since we have $h_{\beta_1}(\mathbf{L}_1) = h_{\beta_2}(\mathbf{L}_2)$ when $\mathbf{L}_1 = (\beta_2/\beta_1)\mathbf{L}_2$. However, no such construction leading to $h_{\beta_1}(\mathbf{L}_1) = h_{\beta_2}(\mathbf{L}_2)$ exists for the other GBF types as shown in (24)–(30), thus the corresponding classes are jointly identifiable. □

For a general choice of $\mathcal{H}$, the following proposition states the sufficient condition for graph identifiability.

**Proposition 2.** *A class of graph systems defined by $\mathcal{L}$ and $\mathcal{H}$ is graph identifiable if $\mathcal{L}$ is a set of CGLs and $\mathcal{H}$ is a set of one-to-one functions.*

*Proof:* Assuming that all $h \in \mathcal{H}$ are one-to-one functions, we need to show that for $\mathbf{L}_1 \neq \mathbf{L}_2$ in $\mathcal{L}$, we have $h(\mathbf{L}_1) \neq h(\mathbf{L}_2)$. Specifically, if $\mathbf{L}_1 = \mathbf{U}\mathbf{\Lambda}_1\mathbf{U}^\intercal$ and $\mathbf{L}_2 = \mathbf{U}\mathbf{\Lambda}_2\mathbf{U}^\intercal$, which have the same GFT, $\mathbf{U}h(\mathbf{\Lambda}_1)\mathbf{U}^\intercal \neq \mathbf{U}h(\mathbf{\Lambda}_2)\mathbf{U}^\intercal$ is satisfied, because $h$ is one-to-one. The proof is obvious when GFTs of $\mathbf{L}_1$ and $\mathbf{L}_2$ are different. □

To derive necessary conditions for graph identifiability, additional set of assumptions are needed. For example, if graphs (i.e., graph Laplacians in $\mathcal{L}$) are assumed to be sparse, the one-to-one requirement on $\mathcal{H}$ may not be necessary. Particularly in [32], identifiability of graphs from their eigenspaces (i.e., from GFTs as in [20]) are investigated, and necessary conditions are introduced for a specific type of identifiability, called *diagonal identifiability*. To the best of our knowledge, there are no other analysis on the identifiability of graphs in the literature. As part of our future work, we will focus on characterizations of necessary and sufficient conditions for joint identifiability under more general choices of $\mathcal{H}$.

## V. Proposed Solution and Optimality Conditions

Algorithm 1 is proposed to solve the GSI problem in (8) for a given sample covariance $\mathbf{S}$ and a selected type of GBF $h_\beta$. After obtaining $\mathbf{U}$ and $\mathbf{\Lambda}_s$ via eigendecomposition of $\mathbf{S}$ and initialization of the parameter $\beta$ (see lines 1 and 2), the algorithm performs three main steps to find the optimal graph Laplacian $\mathbf{L}$ and the filter parameter $\beta$:

**Algorithm 1** Graph System Identification (GSI)

**Input:** Sample covariance $\mathbf{S}$, graph-based filter type $h_\beta$
**Output:** Graph Laplacian $\mathbf{L}$ and $\beta$ filter parameter

1: Obtain $\mathbf{U}$ and $\mathbf{\Lambda}_s$ via eigendecomposition $\mathbf{S} = \mathbf{U}\mathbf{\Lambda}_s\mathbf{U}^\intercal$
2: Initialize parameter $\widehat{\beta}$: For variance or frequency shifting GBFs, apply the initialization methods in Section V-C. For the other GBF types, apply random initialization.
3: **repeat**
4: Prefilter the sample covariance $\mathbf{S}$:

$$\mathbf{S}_{\text{pf}} = (h^{-1}_{\widehat{\beta}}(\mathbf{S}))^{\dagger} = \mathbf{U}(h^{-1}_{\widehat{\beta}}(\mathbf{\Lambda}_s))^{\dagger}\mathbf{U}^\intercal \tag{15}$$

5: Estimate $\mathbf{L}$ from prefiltered data ($\mathbf{S}_{\text{pf}}$):
$\widehat{\mathbf{L}} \leftarrow$ Run the CGL algorithm in [17] to solve (23)
6: Update filter parameter $\widehat{\beta}$ or skip if $\widehat{\beta}$ is optimal:
$\widehat{\beta} \leftarrow$ Apply GBF-specific update in Section V-C
7: **until** convergence has been achieved
8: **return** Graph system parameters $\widehat{\mathbf{L}}$ and $\widehat{\beta}$

1) *Prefiltering step* applies an inverse filtering on the sample covariance $\mathbf{S}$ to reverse the effect of filter $h$ in the graph system. Without the prefiltering step, it may be impossible to effectively recover $\mathbf{L}$ from $\mathbf{S}$, since $\mathbf{\Sigma}^\dagger = h(\mathbf{L})^\dagger$ is generally not a graph Laplacian. The proposed prefiltering allows us to effectively estimate the original eigenvalues of $\mathbf{L}$ (i.e., $\mathbf{\Lambda}_\lambda$) from the prefiltered covariance $\mathbf{S}_{\text{pf}}$ in (15).
2) *Graph Laplacian estimation step* uses the CGL estimation algorithm introduced in [17] to learn $\widehat{\mathbf{L}}$ from the prefiltered covariance $\mathbf{S}_{\text{pf}}$.
3) *Filter parameter selection step* finds the best matching $\beta$ for the graph system. Depending on the type of GBF, we propose different methods for parameter selection.

In the rest of this section, we derive the optimality conditions and discuss optimal prefiltering, graph Laplacian estimation and filter parameter selection in Sections V-A, V-B and V-C, respectively.

### A. Optimal Prefiltering

Let CGL $\mathbf{L}$ and GBF $h$ be the parameters of a graph system so that the covariance matrix of the model in (5) is $\mathbf{\Sigma} = h(\mathbf{L})$. There is a GFT matrix $\mathbf{U}$ jointly diagonalizing $\mathbf{L}$ and $\mathbf{\Sigma}$ such that $\mathbf{L} = \mathbf{U}\mathbf{\Lambda}_\lambda\mathbf{U}^\intercal$ and $\mathbf{\Sigma} = \mathbf{U}\mathbf{\Lambda}_\sigma\mathbf{U}^\intercal$. Under the ideal case that $\mathbf{S} = \mathbf{\Sigma}$ is obtained from an asymptotically large number of samples ($k$), by change of variables, the objective function in (7) becomes

$$\mathcal{J}(\mathbf{U}, h(\mathbf{\Lambda}_\lambda)) = \mathrm{Tr}(\mathbf{U}h(\mathbf{\Lambda}_\lambda)^\dagger\mathbf{\Lambda}_\sigma\mathbf{U}^\intercal) - \log|\mathbf{U}h(\mathbf{\Lambda}_\lambda)^\dagger\mathbf{U}^\intercal| \tag{16}$$

which is simplified using properties of $\mathrm{Tr}(\cdot)$ and $|\cdot|$ as

$$\mathcal{J}(h(\mathbf{\Lambda}_\lambda)) = \mathrm{Tr}(h(\mathbf{\Lambda}_\lambda)^\dagger\mathbf{\Lambda}_\sigma) - \log|h(\mathbf{\Lambda}_\lambda)^\dagger| \tag{17}$$

where the GBF $h$ and the diagonal matrix of graph frequencies $\mathbf{\Lambda}_\lambda$ are unknown, and the diagonal matrix $\mathbf{\Lambda}_\sigma$ is known from data. By letting $\phi_i = h(\lambda_i)^\dagger = (h(\mathbf{\Lambda}_\lambda)^\dagger)_{ii}$ and $\sigma_i^2 = (\mathbf{\Lambda}_\sigma)_{ii}$ for $i = 1, 2, \ldots, n$, we can write (17) as

$$\mathcal{J}(\boldsymbol{\phi}) = \sum_{i=1}^{n} \left(\phi_i \sigma_i^2 - \log(\phi_i)\right), \tag{18}$$

where $\boldsymbol{\phi} = [\phi_1\, \phi_2\, \cdots\, \phi_n]^\mathsf{T}$. In minimization of the convex function (18), the optimal solution satisfies the following necessary and sufficient conditions [33] obtained by taking the derivative of (18) with respect to $\phi_i$ and equating to zero,

$$\frac{\partial \mathcal{J}(\boldsymbol{\phi})}{\partial \phi_i} = \frac{1}{\phi_i} - \sigma_i^2 = 0. \tag{19}$$

By change of variables, the optimality conditions can be stated as

$$h(\lambda_i) = (h(\mathbf{\Lambda}_\lambda))_{ii} = (\mathbf{\Lambda}_\sigma)_{ii} = \sigma_i^2 \quad \forall i. \tag{20}$$

Based on Assumption 1, we can also write (20) as

$$h^{-1}(h(\lambda_i)) = \lambda_i = h^{-1}(\sigma_i^2) \quad \forall i, \tag{21}$$

where $h^{-1}$ is the inverse function of $h$. By using the matrix notation, we can state (21) more compactly as

$$h^{-1}(\mathbf{\Sigma}) = \mathbf{U} h^{-1}(\mathbf{\Lambda}_\sigma)\mathbf{U}^\mathsf{T} = \mathbf{U}\mathbf{\Lambda}_\lambda\mathbf{U}^\mathsf{T} = \mathbf{L}. \tag{22}$$

This condition shows that we can find the optimal Laplacian $\mathbf{L}$ via inverse filtering (inverse prefiltering) $h^{-1}(\mathbf{\Sigma})$. However, in practice, we can only have access to a sample estimate of $\mathbf{\Sigma}$ (i.e., $\mathbf{S}$) obtained from a limited number of data samples ($k$), which is not a good estimator especially when $k$ is small [24]. Thus, computing $h^{-1}(\mathbf{S})$ generally does not lead to a CGL matrix. In order to address this problem, the proposed Algorithm 1 first estimates the prefiltered sample covariance $\mathbf{S}_{\text{pf}}$ as in (15), then employs the CGL estimation algorithm [17] to find the best CGL from $\mathbf{S}_{\text{pf}}$ by minimizing the criterion in (10).

### B. Optimal Graph Laplacian Estimation

For a GBF $h$ (or $h_\beta$) satisfying the optimal prefiltering condition in (21), the GSI problem in (8) can be rewritten as the following graph learning problem,

$$\begin{array}{ll} \underset{\mathbf{L}\succeq 0}{\text{minimize}} & \operatorname{Tr}(\mathbf{L}\mathbf{S}_{\text{pf}}) - \log|\mathbf{L}| + \|\mathbf{L}\odot\mathbf{H}\|_1 \\ \text{subject to} & \mathbf{L}\,\mathbf{1} = \mathbf{0}, \quad (\mathbf{L})_{ij} \le 0 \quad i \ne j \end{array} \tag{23}$$

where $\mathbf{S}_{\text{pf}} = (h^{-1}(\mathbf{S}))^\dagger$ is obtained by prefiltering the empirical covariance $\mathbf{S}$. This problem is discussed in detail in our previous work [17] where we have derived the optimality conditions for (23) and developed the CGL estimation algorithm to solve it.

### C. Filter Parameter Selection

Based on the optimality condition in (21), we propose different methods to identify the parameter $\beta$ for GBFs in Table I. Specifically, optimal parameter initializations for variance and frequency shifting GBFs are derived, and a line search method is proposed for $\beta$-hop localized GBF. Since graph systems with exponential decay and frequency scaling GBFs are not jointly identifiable (as discussed in Section IV-D), we cannot identify $\beta$ optimally. Yet, we show that graph Laplacian matrices can be identified up to a constant factor, which depends on $\beta$.

**Initialization for Variance/Frequency Shifting Filters.** For both variance and frequency shifting GBFs, the optimal $\beta$ is found by calculating $\sigma_1^2 = \mathbf{u}_1{}^\mathsf{T}\mathbf{\Sigma}\mathbf{u}_1$ where $\mathbf{u}_1$ is the eigenvector associated with the zero eigenvalue of the Laplacian $((\mathbf{\Lambda}_\lambda)_{11} = \lambda_1 = 0)$. Specifically, by using the optimality condition in (21),

- if $h_\beta(\lambda)$ is a variance shifting filter, we get
$$h_\beta(\lambda_1) = \sigma_1^2 = \mathbf{u}_1{}^\mathsf{T}\mathbf{\Sigma}\mathbf{u}_1 = \lambda_1^\dagger + \beta, \tag{24}$$
since $\lambda_1 = \lambda_1^\dagger = 0$, the optimal $\beta$ satisfies
$$\beta = \mathbf{u}_1{}^\mathsf{T}\mathbf{\Sigma}\mathbf{u}_1 = \sigma_1^2. \tag{25}$$
- if $h_\beta(\lambda)$ is a frequency shifting filter, we obtain
$$h_\beta(\lambda_1) = \sigma_1^2 = \mathbf{u}_1{}^\mathsf{T}\mathbf{\Sigma}\mathbf{u}_1 = 1/(\lambda_1 + \beta) = 1/\beta, \tag{26}$$
so the optimal $\beta$ satisfies
$$\beta = 1/(\mathbf{u}_1{}^\mathsf{T}\mathbf{\Sigma}\mathbf{u}_1) = 1/\sigma_1^2. \tag{27}$$

Since the optimal $\beta$ can be directly estimated from the sample covariance $\mathbf{S}$ as calculated from $\mathbf{\Sigma}$ in (25) and (27), Algorithm 1 uses the optimized initial $\beta$ (in line 2) for prefiltering, and then estimates $\mathbf{L}$, so that the filter parameter update (line 6) is skipped for graph systems with frequency and variance shifting filters.

**Exponential Decay and Frequency Scaling Filters.** Suppose that $\widehat{\mathbf{L}}$ is obtained by inverse prefiltering of $\mathbf{\Sigma}$ with $\widehat{\beta}$, that is formally, $\widehat{\mathbf{L}} = h_{\widehat{\beta}}^{-1}(\mathbf{\Sigma})$.

- If $h_\beta(\lambda)$ is a frequency scaling filter, we have
$$h_{\widehat{\beta}}^{-1}(\sigma_i^2) : \lambda_1 = \sigma_1^2 = 0, \;\; \frac{\beta}{\widehat{\beta}}\lambda_i = \frac{1}{\widehat{\beta}\sigma_i^2} \quad i = 2, \ldots, n. \tag{28}$$
- If $h_\beta(\lambda)$ is an exponential decay filter, we have
$$h_{\widehat{\beta}}^{-1}(\sigma_i^2) = -\frac{\log(\sigma_i^2)}{\widehat{\beta}} = -\frac{\log(\exp(-\beta\lambda_i))}{\widehat{\beta}} = \frac{\beta}{\widehat{\beta}}\lambda_i. \tag{29}$$

Based on (28) and (29), for any selected $\widehat{\beta}$, the resulting matrix is a CGL satisfying $\widehat{\mathbf{L}} = (\beta/\widehat{\beta})\mathbf{L}$ where $\mathbf{L}$ and $\beta$ denote the original graph system parameters. Since the inverse of a different GBF (i.e., $h_{\widehat{\beta}}^{-1}$ with $\beta \ne \widehat{\beta}$) leads to a CGL for any $\beta$, Algorithm 1 can only find the optimal CGL matrix $\mathbf{L}$ up to a constant factor $\beta/\widehat{\beta}$. In practice, the parameter $\widehat{\beta}$ can be tuned so that the desired normalization (scaling factor) for $\mathbf{L}$ is achieved.

**$\beta$-hop Localized Filter.** For estimation of the optimal hop count $\beta$ in Algorithm 1, inverse prefiltering with $\widehat{\beta}$ gives,

$$h_{\widehat{\beta}}^{-1}(\sigma_i^2) : \lambda_1 = \sigma_1^2 = 0, \;\; \lambda_i^{\beta/\widehat{\beta}} = \left(\frac{1}{\sigma_i^2}\right)^{1/\widehat{\beta}} \quad i = 2, \ldots, n. \tag{30}$$

Since this requires the graph learning step to estimate $\mathbf{L}^{\beta/\widehat{\beta}}$, which is not a graph Laplacian in general, Algorithm 1 cannot guarantee optimal graph system identification unless $\beta = \widehat{\beta}$. In order to find the optimal $\beta$, we perform a line search for given range of integers optimizing the following:

$$\widehat{\beta} = \underset{\widehat{\beta}\in\mathbb{N}}{\operatorname{argmin}} \|h_{\widehat{\beta}}(\widehat{\mathbf{L}}) - \mathbf{S}\|_F. \tag{31}$$

## VI. RESULTS

### A. Graph Learning from Diffusion Signals/Data

We evaluate the performance of our proposed graph system identification method (GSI or Algorithm 1) by benchmarking against the current state-of-the-art approaches proposed for learning graph from smooth signals (GLS) [18], [19] as well as the graph topology inference (GTI) in [20]. The proposed GSI is also compared against the CGL estimation algorithm, CGL [17] (i.e., using the CGL estimation algorithm without prefiltering), and the inverse prefiltering (IPF) approach, which estimates a graph Laplacian matrix by inverting the prefiltered covariance, $\mathbf{S}_{\text{pf}}$ in (15), so that $\widehat{\mathbf{L}} = h_\beta^{-1}(\mathbf{S}) = \mathbf{S}_{\text{pf}}^\dagger$. For this purpose, we generate several artificial datasets based on the signal model in (5), defined by a graph Laplacian ($\mathbf{L}$) and a GBF ($h_\beta$) where the dataset entries are generated by random sampling from the distribution $\mathsf{N}(\mathbf{0}, h_\beta(\mathbf{L}))$. Then, the generated data is used in the proposed and benchmark algorithms to recover the corresponding graph Laplacian $\mathbf{L}$. We repeat our experiments for different $\mathbf{L}$ and $h_\beta$ where graphs are constructed by using three different graph connectivity models:

- Grid graph, $\mathcal{G}_{\text{grid}}^{(n)}$, consisting $n$ vertices attached to their four nearest neighbors (except the vertices at boundaries).
- Random Erdos-Renyi graph, $\mathcal{G}_{\text{ER}}^{(n,p)}$, with $n$ vertices attached to other vertices with probability $p = 0.2$.
- Random modular graph (also known as stochastic block model), $\mathcal{G}_{\text{M}}^{(n,p_1,p_2)}$ with $n$ vertices and four modules (subgraphs) where the vertex attachment probabilities across modules and within modules are $p_1 = 0.1$ and $p_2 = 0.2$, respectively.

Then, the edge weights of a graph are randomly selected from the uniform distribution $\mathsf{U}(0.1, 3)$, on the interval $[0.1, 3]$. For each $\mathbf{L}$ and $h_\beta$ pair, we perform Monte-Carlo simulations to test average performance of proposed and benchmark methods with varying number of data samples ($k$) and fixed number of vertices ($n = 36$)[9]. To measure the estimation performance, we use the following two metrics:

$$\mathsf{RE}(\widehat{\mathbf{L}}, \mathbf{L}^*) = \frac{\|\widehat{\mathbf{L}} - \mathbf{L}^*\|_F}{\|\mathbf{L}^*\|_F} \tag{32}$$

which is the relative error between the ground truth graph ($\mathbf{L}^*$) and estimated graph parameters ($\widehat{\mathbf{L}}$), and

$$\mathsf{FS}(\widehat{\mathbf{L}}, \mathbf{L}^*) = \frac{2\,\mathsf{tp}}{2\,\mathsf{tp} + \mathsf{fn} + \mathsf{fp}} \tag{33}$$

is the F-score metric (commonly used metric to evaluate binary classification performance) calculated based on true-positive (tp), false-positive (fp) and false-negative (fn) detection of graph edges in estimated $\widehat{\mathbf{L}}$ with respect to the ground truth edges in $\mathbf{L}^*$. F-score takes values between 0 and 1, where the value 1 means perfect classification.

In our experiments, for the proposed GSI, the regularization parameter $\alpha$ in (23) is selected from the following set:

$$\{0\} \cup \{0.75^r (s_{\max}\sqrt{\log(n)/k}) \,|\, r = 1, 2, 3, \dots, 14\}, \tag{34}$$

where $s_{\max} = \max_{i \neq j} |(\mathbf{S})_{ij}|$ is the maximum off-diagonal entry of $\mathbf{S}$ in absolute value, and the scaling term $\sqrt{\log(n)/k}$ is used for adjusting the regularization according to $k$ and $n$ as suggested in [24], [35]. Monte-Carlo simulations are performed for each proposed/baseline method, by successively solving the associated problem with different regularization parameters to find the best regularization that minimizes RE. The corresponding graph estimate is also used to calculate FS. For all baseline methods [18]–[20], the required parameters are selected by fine tuning. Since CGL [17], GLS [18], [19] and GTI [20] approaches generally result in severely biased solutions with respect to the ground truth $\mathbf{L}^*$ (based on our observations from the experiments), RE values are calculated after normalizing the estimated solution $\mathbf{L}$ as $\widehat{\mathbf{L}} = (\text{Tr}(\mathbf{L}^*)/\text{Tr}(\mathbf{L}))\mathbf{L}$. Note that, this normalization also resolves the ambiguity in identification of graph systems with exponential decay and frequency scaling filters up to a scale factor (discussed in Section V-C).

Figures 3 and 4 depict the performance of different methods applied for estimating graphs from signals modeled based on exponential decay filters (diffusion kernels) and $\beta$-hop localized filters. As shown in Figs. 3 and 4, the proposed GSI significantly outperforms all baseline methods, including the state-of-the-art GLS [18], [19] and GTI [20], in terms of average RE and FS metrics. The performance difference between GSI and CGL [17] demonstrates the impact of the prefiltering step, which substantially improves the graph learning accuracy. Similarly, the performance gap between GSI and IPF shows the advantage of Algorithm 1 compared to the direct prefiltering of input covariance ($\mathbf{S}$) as in (22), where GSI provide better graph estimation especially when number of data samples (i.e., $k/n$) is small. Besides, Figs. 5 and 6 illustrate two examples from the experiments with grid graphs for the case of $k/n = 30$, where the proposed GSI constructs graphs that are the most similar to the ground truth ($\mathbf{L}^*$).

### B. Graph Learning from Variance/Frequency Shifted Signals

In this subsection, we compare the CGL estimation performance of GSI, CGL [17] and SCGL [30] methods from signals modeled based on variance and frequency shifting GBFs. As discussed in Section IV-C, the covariance matrices for signals modeled based on these GBFs with parameter $\beta$ are

- $\mathbf{\Sigma} = \mathbf{L}^\dagger + \beta\mathbf{I}$ for variance shifting,
- $\mathbf{\Sigma} = (\mathbf{L} + \beta\mathbf{I})^\dagger$ for frequency shifting.

where $\mathbf{L}$ denotes the associated combinatorial Laplacian.

In our experiments, we construct 10 random Erdos-Renyi graphs ($\mathcal{G}_{\text{ER}}^{(n,p)}$) with $n = 36$ vertices and $p = 0.2$, then generate $\mathbf{\Sigma}$ for each GBF by varying $\beta$ between 0 and 1. To evaluate the effect of $\beta$ only, we use actual covariance matrices instead of sample covariances as input to the algorithms. So, GSI, CGL and SCGL estimate a graph Laplacian $\mathbf{L}$ from $\mathbf{\Sigma}$. The average RE results are presented in Tables IV and V for various $\beta$.

Table IV shows that the proposed GSI significantly outperforms CGL for $\beta > 0$, and the average RE difference increases as $\beta$ gets larger. This is because the variance shifting GBF leads to the noisy signal model with the covariance in (11) where $\beta$ represents the variance of the noise ($\sigma^2$), and the prefiltering step allows GSI to perfectly estimate the parameter $\beta$ from $\mathbf{\Sigma}$ by using (25) so that the covariance is prefiltered as in (15) based on the optimal $\beta$. The prefiltering step can also be considered as a *denoising operation* (reversing the effect of variance shifting GBFs) on the signal covariance before the graph estimation step, while CGL works with noisy (i.e., shifted) covariances, which diminish the CGL estimation performance. For $\beta = 0$ (i.e., $\mathbf{\Sigma}$ is noise-free), the problem (8) reduces to the CGL estimation problem in [17], so both GSI and CGL lead to the same average RE.

[9]Methods are evaluated on small graphs (with $n = 36$ vertices), since GTI [20] is implemented using CVX [34] and do not currently have an efficient and scalable implementation. The proposed and the other benchmark methods are more efficient and can support larger graphs.

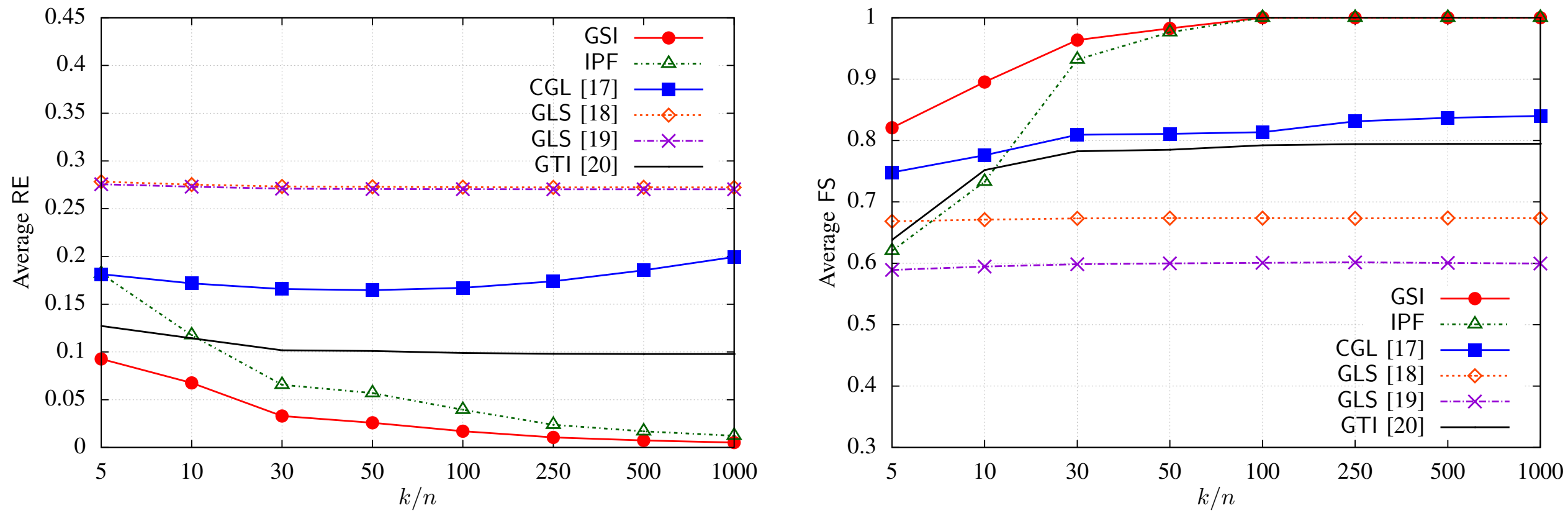

Fig. 3. Average RE and FS results for graph estimation from signals modeled based on exponential decay GBFs tested with $\beta = \{0.5, 0.75\}$ on 10 different grid, Erdos-Renyi and modular graphs (30 graphs in total). The proposed GSI outperforms all baseline methods in terms of both RE and FS.

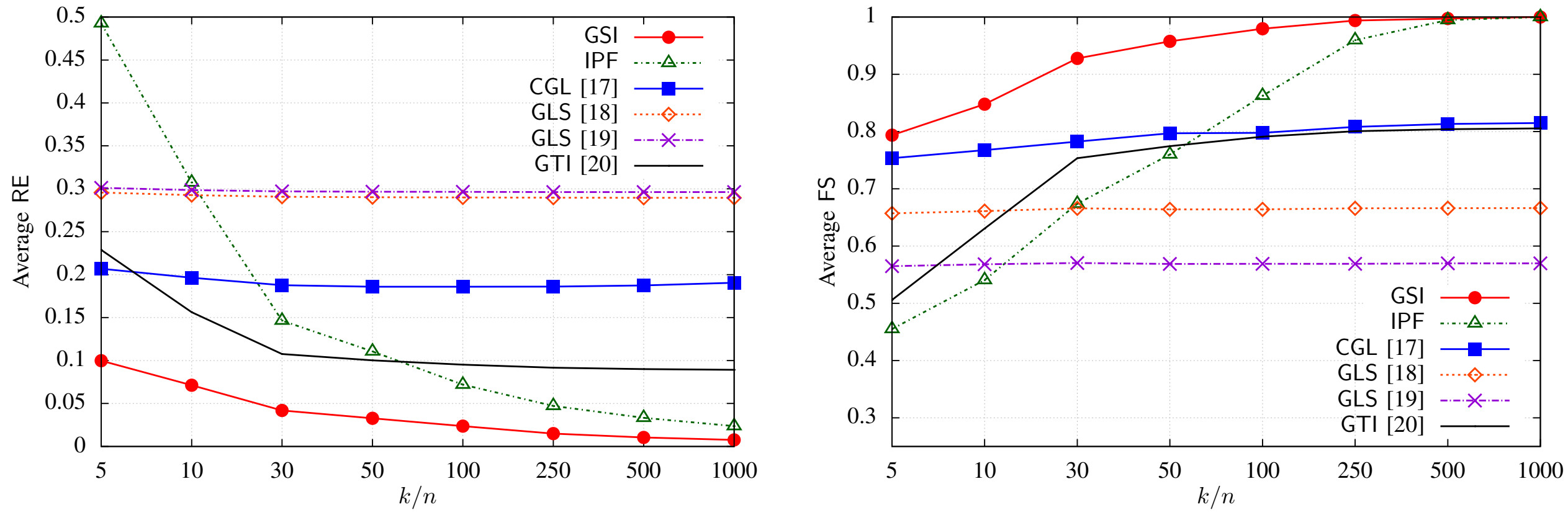

Fig. 4. Average RE and FS results for graph estimation from signals modeled based on $\beta$-hop localized GBFs tested with $\beta = \{2, 3\}$ on 10 different grid, Erdos-Renyi and modular graphs (30 graphs in total). The proposed GSI outperforms all baseline methods in terms of both RE and FS.

TABLE IV
AVERAGE RELATIVE ERRORS FOR VARIANCE SHIFTING GBF

| | **Filter parameter** ($\beta$) | | | | | |
|---|---|---|---|---|---|---|
| **Method** | 0 | 0.1 | 0.3 | 0.5 | 0.7 | 0.9 |
| CGL [17] | $2\times10^{-4}$ | 0.60 | 0.79 | 0.85 | 0.88 | 0.89 |
| GSI | $2 \times 10^{-4}$ | | | | | |

TABLE V
AVERAGE RELATIVE ERRORS FOR FREQUENCY SHIFTING GBF

| | **Filter parameter** ($\beta$) | | | |
|---|---|---|---|---|
| **Method** | 0 | 0.1 | 0.5 | 0.9 |
| SCGL [30] | 0.2354 | $6.7\times10^{-4}$ | $6.6\times10^{-4}$ | $6.2\times10^{-4}$ |
| GSI | $1.6 \times 10^{-4}$ | | | |

For the frequency shifting GBFs with $\beta > 0$, GSI performs slightly better than SCGL, since SCGL is implemented using a general purpose solver CVX [34], which generally produces less accurate solutions compared to our algorithm. Moreover, our algorithm is approximately 90 times faster than SCGL on average, and significantly outperforms SCGL for $\beta = 0$, since SCGL method is developed for shifted covariance matrices (i.e., $\mathbf{L} + \beta\mathbf{I}$) where $\beta$ needs to be strictly positive.

### C. Illustrative Results on Temperature Data

In this experiment, we apply our proposed method on a real (climate) dataset[10] consisting of air temperature measurements [36]. We specifically use the average daily temperature measurements collected from 45 states in the US over 16 years (2000-2015), so that in total there are $k = 5844$ samples for each of the $n = 45$ states. Fig. 7 shows samples of average temperature signals, which are spatially smooth across different states. Also, *the Rocky Mountains* region has lower average temperature values as compared to the other regions in the western US[11].

[10]NCEP Reanalysis data provided by the NOAA/OAR/ESRL PSD, Boulder, Colorado, USA, from their website at http://www.esrl.noaa.gov/psd/

[11]The Rocky Mountains cross through the states of Idaho, Montana, Wyoming, Utah, Colorado and New Mexico. For example, in Fig. 7b, the areas with temperature values between 0 and 10 degrees Celsius (colored in green) correspond to the Rocky Mountains region approximately.

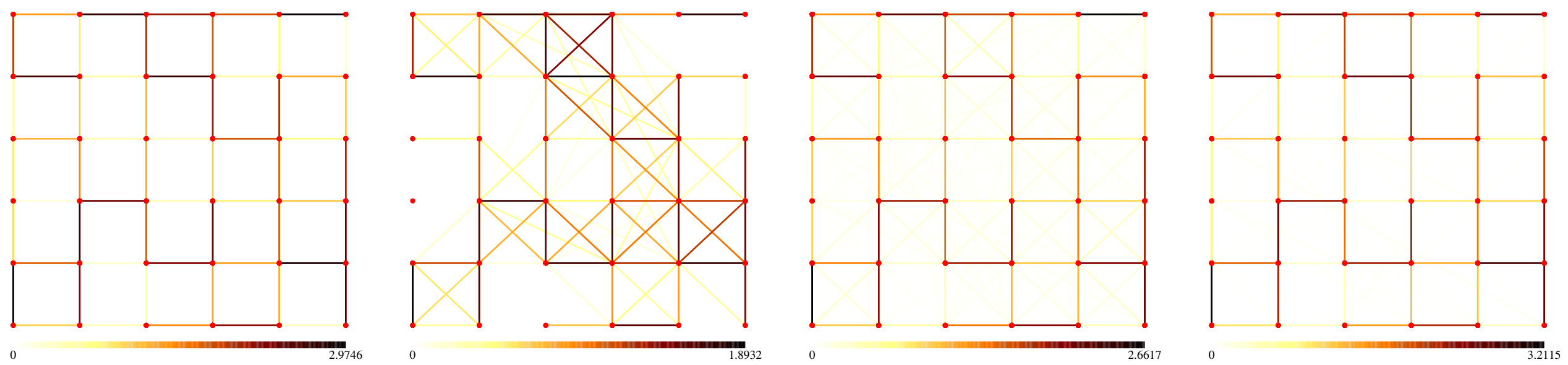

(a) The ground truth CGL ($\mathbf{L}^*$) (b) GLS [18] (RE,FS) = (0.4019,0.6172) (c) GTI [20] (RE,FS) = (0.1201,0.7229) (d) Proposed GSI (RE,FS) = (0.0340,0.9917)

Fig. 5. A sample illustration of graph estimation results (for $k/n = 30$) from signals modeled using the exponential decay GBF with $\beta = 0.75$ and $\mathbf{L}^*$ is derived from the grid graph in (a). The edge weights are color coded where darker colors indicate larger weights. The proposed GSI leads to the graph that is the most similar to the ground truth.

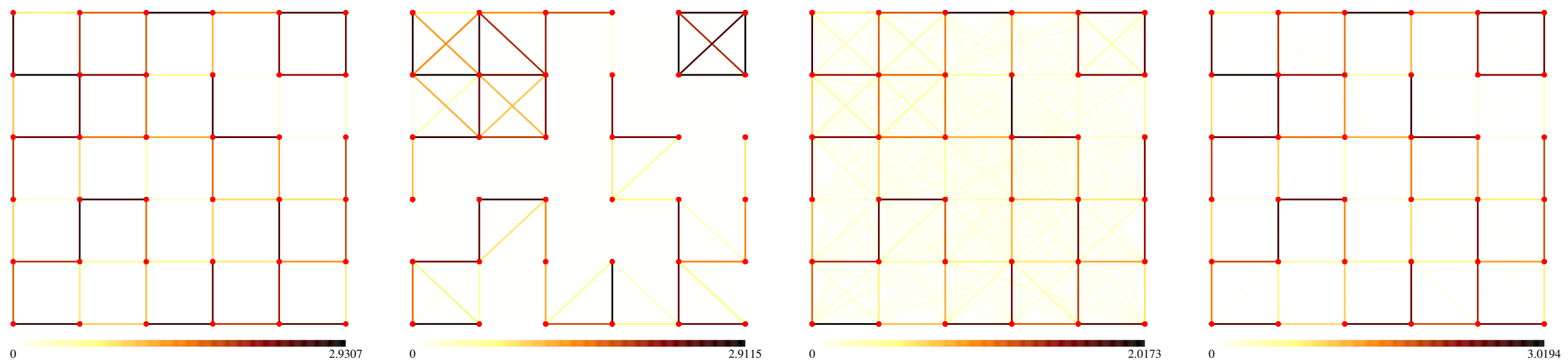

(a) The ground truth CGL ($\mathbf{L}^*$) (b) GLS [18] (RE,FS) = (0.4045,0.6890) (c) GTI [20] (RE,FS) = (0.2118,0.6030) (d) Proposed GSI (RE,FS) = (0.0274,0.9833)

Fig. 6. A sample illustration of graph estimation results (for $k/n = 30$) from signals modeled using the $\beta$-hop localized GBF with $\beta = 2$ and $\mathbf{L}^*$ is derived from the grid graph in (a). The edge weights are color coded where darker colors indicate larger weights. The proposed GSI leads to the graph that is the most similar to the ground truth.

The goal of this experiment is to learn graphs (with 45 vertices) associated with exponential decay and $\beta$-hop localized filters from temperature data and investigate the effect of filter type and parameter $\beta$ on the resulting graphs, representing the similarity of temperature conditions between the 45 states. For modeling temperature signals, a diffusion kernel is a good candidate, because it is a fundamental solution of the *heat equation*[12], which describes the distribution of heat in a physical environment [37].

Figure 8 illustrates the graphs estimated using the GSI method without $\ell_1$-regularization (i.e., $\mathbf{H}$ in (8) is set to the zero matrix). As shown in the figure, larger edge weights are assigned between vertices (i.e., states) that are closer to each other in general, since temperature values are mostly similar between states within close proximity. However, the distance between states is obviously not the only factor effecting the similarity of temperature values. For example, in Figs. 8b–8f, the weights are considerably small between the states in the Rocky Mountains region and their neighboring states in the east (e.g., Nebraska and Kansas) due to the large differences in altitude. Note also that different choices of GBFs can lead to substantially different similarity graphs. Especially for the $\beta$-hop localized GBF with $\beta = 1$ (corresponding to the CGL method), the resulting graph is significantly different than the results in Figs. 8b and 8c, since the 1-hop localized filter does not lead to a diffusion model. The graphs associated with exponential decay GBF leads to sparser graphs, better revealing the structure of the signal, compared to the graphs in Figs. 8a–8c. The structure of the graphs in Figs. 8d–8f are similar for different $\beta$ because of the relation in (29) for the exponential decay filter. For example, increasing $\beta$ from 0.25 to 0.5 approximately halves edge weights, as shown in Figs. 8d and 8e. Besides, the distribution of edge weights for $\beta$-hop localized GBFs becomes more similar to the ones in Figs. 8d–8f as $\beta$ gets larger.

[12]This is the reason that diffusion kernels are also known as heat kernels in the literature.

## VII. Conclusion

In this paper, we have introduced a novel graph-based modeling framework that (i) formulates the modeling problem as the graph system identification from signals/data and (ii) proposes an alternating optimization algorithm iteratively solving for a graph and a GBF. At each iteration of the algorithm, a prefiltering operation, defined by the estimated GBF, is applied on the observed signals, and then a graph is estimated from prefiltered signals. The experimental results have demonstrated that the proposed algorithm outperforms the existing methods on modeling smooth signals and learning diffusion-based models [18]–[20] in terms of graph estimation accuracy.

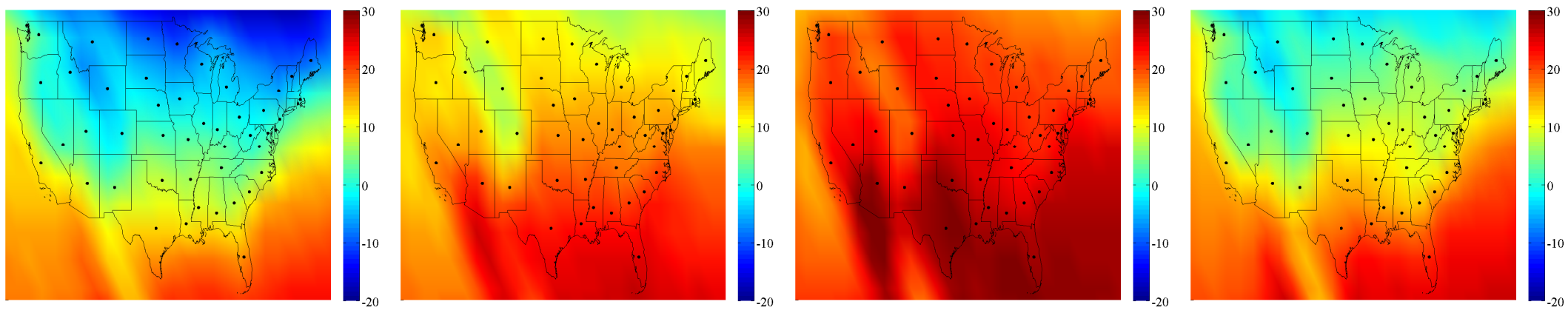

(a) 45th day (Winter) (b) 135th day (Spring) (c) 225th day (Summer) (d) 315th day (Autumn)

Fig. 7. Average air temperatures (in degree Celsius) for (a) 45th, (b) 135th, (c) 225th and (d) 315th days over 16 years (2000-2015). Black dots denote 45 states.

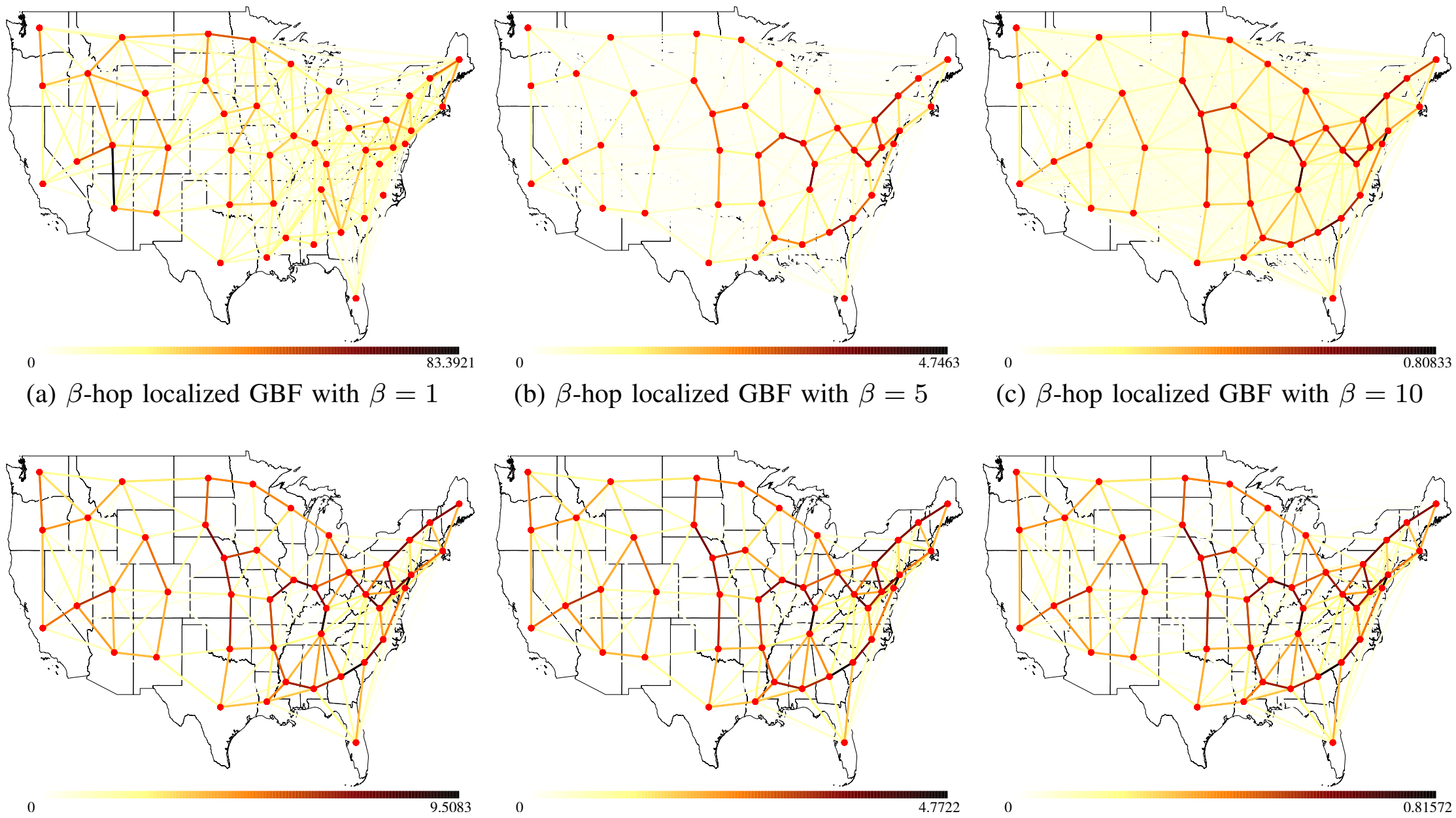

(a) $\beta$-hop localized GBF with $\beta = 1$ (b) $\beta$-hop localized GBF with $\beta = 5$ (c) $\beta$-hop localized GBF with $\beta = 10$

(d) Exponential decay with $\beta = 0.25$ (e) Exponential decay with $\beta = 0.5$ (f) Exponential decay with $\beta = 3$

Fig. 8. Graphs learned from temperature data using the GSI method with exponential decay and $\beta$-hop localized GBFs for fixed $\beta$ parameters where no regularization is applied (i.e., $\mathbf{H}$ is set to the zero matrix). The edge weights are color coded so that darker colors represent larger weights (i.e., more similarities). The graphs associated with exponential decay GBF leads to sparser structures compared to the graphs corresponding to $\beta$-hop localized GBFs.

The proposed framework supports various types of GBFs (in Table I) including the diffusion (heat) kernels as special cases which are widely used in many applications [4], [5], [27]. Our future work focuses on the extensions of our algorithm for joint identification of graphs and polynomial filters (i.e., estimation of polynomials of graph Laplacians), which can provide more degrees of freedom in designing filters than the GBFs in Table I. Also, data-oriented applications of the proposed modeling framework is considered as part of our future work.

## Appendix A
## Derivation of Diffusion Kernels

Suppose that the random process is obtained by diffusion of the initial random vector $\mathbf{x}(0)$ whose entries are independent, zero-mean random variables, denoted as $(\mathbf{x}(0))_i$ for $i=1,2,\dots,n$, with variance $\sigma^2$, the random diffusion over a graph $\mathcal{G}$ is formulated recursively as

$$\mathbf{x}(t+1) = \mathbf{x}(t) - r\mathbf{L}\mathbf{x}(t) = (\mathbf{I} - r\mathbf{L})\mathbf{x}(t) \quad t \in \{0,1,\dots\} \tag{35}$$

where $t$ represents the time index, $\mathbf{L}$ is the graph Laplacian of $\mathcal{G}$ and the real number $r \in (0,1)$ is the diffusion rate of the process. On $i$-th vertex of the graph $\mathcal{G}$, we can write the random process in (35) as

$$(\mathbf{x}(t+1))_i = (\mathbf{x}(t))_i + r \sum_{j \in \mathcal{N}_i} (\mathbf{W})_{ij} \left((\mathbf{x}(t))_j - (\mathbf{x}(t))_i\right) \tag{36}$$

for all $i$ where $\mathbf{W}$ is the adjacency matrix of the graph $\mathcal{G}$, and $\mathcal{N}_i$ is the set of vertex indices neighboring $i$-th vertex $(v_i)$. More compactly, (35) can be written as

$$\mathbf{x}(t) = \mathbf{G}(t)\mathbf{x}(0) \text{ where } \mathbf{G}(t) = (\mathbf{I} - r\mathbf{L})^t. \tag{37}$$

The covariance of $\mathbf{x}(t)$ is

$$\begin{aligned}(\mathbf{\Sigma}(t))_{ij} &= \mathsf{E}\left[(\mathbf{G}(t)\mathbf{x}(0))_i(\mathbf{G}(t)\mathbf{x}(0))_j\right] \\ &= \mathsf{E}\left[\left(\sum_{k=1}^{n}(\mathbf{G}(t))_{ik}(\mathbf{x}(0))_k\right)\left(\sum_{l=1}^{n}(\mathbf{G}(t))_{jl}(\mathbf{x}(0))_l\right)\right],\end{aligned} \tag{38}$$

which simplifies by using independence of $\mathbf{x}(0)$ as

$$(\mathbf{\Sigma}(t))_{ij} = \sigma^2 \sum_{k=1}^{n}((\mathbf{G}(t))_{ik}(\mathbf{G}(t))_{kj}) = \sigma^2(\mathbf{G}(t)^2)_{ij}. \tag{39}$$

Therefore, the covariance matrix leads to

$$\boldsymbol{\Sigma}(t) = \mathbf{G}(t)^2 = \sigma^2 \left(\mathbf{I} - r\mathbf{L}\right)^{2t}. \quad (40)$$

To adjust the time resolution of the process, we replace $t$ with $t/\Delta t$ and $r$ with $r\Delta t$ in $\mathbf{G}(t)$ so that

$$\mathbf{G}(t) = \sigma \left(\mathbf{I} - \frac{r\mathbf{L}}{1/\Delta t}\right)^{t/\Delta t}. \quad (41)$$

For arbitrarily small $\Delta t$, $\mathbf{G}(t)$ converges to a matrix exponential function of $\mathbf{L}$,

$$\widetilde{\boldsymbol{\Sigma}}(t) = \sigma^2 \lim_{\Delta t \to 0} \left(\mathbf{I} - \frac{r\mathbf{L}}{1/\Delta t}\right)^{2t/\Delta t} = \sigma^2 \exp(-2rt\mathbf{L}) \quad (42)$$

which is equivalent to exponential decay filtering operation in Table I with $\beta(t) = -2rt$. For arbitrarily large $t$ (i.e., when the diffusion reaches steady-state), the covariance is

$$\lim_{t\to\infty} \widetilde{\boldsymbol{\Sigma}}(t) = \lim_{t\to\infty} \sum_{i=1}^{n} \exp(-2rt\lambda_i)\mathbf{u}_i {\mathbf{u}_i}^{\mathsf{T}} = \frac{\sigma^2}{n}\mathbf{1}\mathbf{1}^{\mathsf{T}}, \quad (43)$$

whose all entries have the same value, since we have $\lambda_1 = 0$ for CGLs. Thus, the statistical properties eventually become spatially flat across all vertices as intuitively expected for a diffusion process. Yet, for nonsingular graph Laplacians (e.g., generalized graph Laplacians [17], [26]), the process behaves differently so that the signal energy vanishes, since the above limit converges to the $n \times n$ zero matrix as $t$ gets larger.